# Evaluating Cultural Knowledge Processing in Large Language Models: A Cognitive Benchmarking Framework Integrating Retrieval-Augmented Generation

Hung-Shin Lee[1], Chen-Chi Chang[2,*] Ching-Yuan Chen[2] and Yun-Hsiang Hsu[2]

[1]United Link Co., Ltd.

[2]Department of Cultural Creativity and Digital Marketing, National United University

[*]Correspondence: kiwi@gm.nuu.edu.tw

## ABSTRACT

**Design/methodology/approach**

This study proposes a cognitive benchmarking framework to evaluate how large language models (LLMs) process and apply culturally specific knowledge. The framework integrates Bloom's Taxonomy with Retrieval-Augmented Generation (RAG) to assess model performance across six hierarchical cognitive domains: Remembering, Understanding, Applying, Analyzing, Evaluating, and Creating. Using a curated Taiwanese Hakka digital cultural archive as the primary testbed, the evaluation measures LLM-generated responses' semantic accuracy and cultural relevance.

**Purpose**

This research evaluates how effectively LLMs represent and generate minority cultural knowledge, specifically Taiwanese Hakka culture. To address this, the study proposes a structured and replicable evaluation framework integrating Bloom's Taxonomy and RAG. The research is guided by the following questions: (1) How do LLMs perform across different cognitive domains when processing Hakka cultural content? (2) To what extent does the integration of RAG enhance the accuracy and contextual appropriateness of LLM outputs? (3) How do different LLM architectures compare in their ability to recall, analyze, and creatively synthesize culturally grounded information?

**Findings**

The evaluation results indicate that LLMs augmented with RAG exhibit marked improvements over baseline models in the cognitive domains of Remembering, Understanding, and Analyzing. These enhancements are particularly evident in tasks requiring factual accuracy, contextual relevance, and semantic precision, underscoring

RAG's effectiveness in addressing the knowledge sparsity typically observed in underrepresented cultural datasets. However, a notable limitation persists across all models including those equipped with RAG in the domain of Creating. This suggests that while retrieval mechanisms bolster the reproduction and comprehension of cultural knowledge, they do not yet sufficiently support culturally nuanced generative synthesis.

**Originality/value**

This study introduces a novel evaluation framework integrating cognitive domain benchmarks with RAG-enhanced LLMs to assess cultural knowledge processing. The research advances culturally grounded AI systems and digital archival quality by empirically demonstrating RAG's impact on improving factual accuracy in lower and mid-level tasks. The findings affirm the strategic value of retrieval integration for enhancing representational fidelity in cultural AI applications, while also highlighting the need for future research into hybrid architectures that combine external grounding with culturally adaptive generation strategies.

# 1. Introduction

Recent research underscores the growing importance of artificial intelligence (AI) in developing culturally aware information systems. Large language models (LLMs) have emerged as powerful tools that can autonomously process, generate, and disseminate cultural knowledge. These capabilities open new possibilities for cross-cultural understanding and heritage preservation. However, despite their technical sophistication, LLMs frequently exhibit limitations and biases, especially when applied to minority cultures with scarce linguistic and cultural resources (C. Gao et al., 2023) . These biases can compromise the fairness, accuracy, and authenticity of cultural representation, ultimately affecting the inclusivity and effectiveness of digital archival systems.

One critical risk is the "AI hallucinations," wherein LLMs generate inaccurate or misleading content (Maleki, Padmanabhan, & Dutta, 2024). For underrepresented cultural communities, such as the Taiwanese Hakka, hallucinated outputs may distort key cultural meanings, reinforce stereotypes, or erase nuanced historical and linguistic contexts. These challenges highlight the urgent need for structured evaluation frameworks integrating human cognitive theories with AI architectures to ensure that AI-generated content reflects cultural accuracy, contextual integrity, and epistemic responsibility.

This study addresses these concerns by proposing a structured and replicable framework that integrates Bloom's Taxonomy and Retrieval-Augmented Generation (RAG). Bloom's Taxonomy, developed as a pedagogical tool (Bloom, 1956; Furst, 1981; Seddon, 1978), provides a hierarchical classification of cognitive processes ranging from factual recall (Remembering) to creative synthesis (Creating), allowing for the systematic evaluation of model outputs across varying depths of cultural understanding. Conversely, RAG enhances LLMs' capabilities by dynamically retrieving external information during inference, thereby improving accuracy and grounding generated responses in contextually relevant knowledge (Lewis et al., 2020).

Representation in generative AI involves more than surface-level inclusion of cultural facts; it requires the accurate, context-sensitive reproduction of a group's knowledge systems and lived experiences. Many minority cultures are underrepresented in pretraining corpora, which results in shallow or reductive portrayals of their traditions, language, and heritage. By integrating Bloom's Taxonomy, this study introduces a hybrid evaluation approach that enables multidimensional assessment of how LLMs engage with minority cultural content.

Taking the Taiwanese Hakka cultural archive as a case study, this research examines how different LLM architectures perform across Bloom's six domains and investigates the role of RAG in enhancing the cultural fidelity of AI-generated outputs. Specifically, the study is guided by the following research questions: (1) How do LLMs perform across different cognitive domains when processing Hakka cultural content? (2) To what extent does the integration of RAG enhance the accuracy and contextual appropriateness of LLM outputs? (3) How do different LLM architectures compare in their ability to recall, analyze, and creatively synthesize culturally grounded information?

Through this interdisciplinary inquiry, the study aims to establish a foundational approach for evaluating and improving the cultural competence of AI systems in underrepresented knowledge domains. It offers a theoretical contribution to the responsible use of Bloom's framework and practical implications for designing inclusive, context-aware cultural archives. These findings have broader relevance for researchers, designers, and policymakers committed to equitable digital heritage representation and culturally sensitive AI governance.

## 2. Literature Review

### 2.1. Cognitive Domain

The cognitive domain, as outlined by Bloom in 1956, encompasses the acquisition of knowledge and the cultivation of intellectual skills. It involves the recall or recognition of specific facts, procedural patterns, and concepts that contribute to the development of cognitive abilities and intellectual competencies. Bloom's Taxonomy originated in 1956, introduced by the American educational psychologist Benjamin Bloom and his colleagues (Bloom, 1956). It was initially developed to establish a systematic framework within the field of education, providing educators with a clear reference standard for designing curricula and assessing student learning outcomes. Since its inception, Bloom's Taxonomy has been widely adopted across all levels of education, becoming an essential tool for instructional design and student assessment. It helps educators systematically design curricula and evaluate student performance across various cognitive levels. With the rapid advancement of digital technologies, Bloom's Taxonomy has gradually adapted to the demands of digital learning environments, serving as the theoretical foundation for knowledge assessment in digital learning (Amin & Mirza, 2020).

Recent studies affirm that Bloom's Taxonomy remains a valuable framework for structuring AI-based cognitive assessment in educational contexts. AI-driven tools have demonstrated the capacity to generate and evaluate questions aligned with the taxonomy's six cognitive levels, enhancing assessment quality and operational efficiency (Yaacoub, Da-Rugna, & Assaghir, 2025). Empirical evidence indicates that embedding Bloom's Taxonomy within AI-supported learning environments can foster higher-order thinking skills, although difficulties persist in transferring and applying knowledge (Elim, 2024; Hui, 2025). Models such as GPT-3.5 and RoBERTa can produce and grade questions across cognitive levels, albeit with notable performance variation (Gani, Ayyasamy, Sangodiah, & Fui, 2023; Hwang, Challagundla, Alomair, Chen, & Choa, 2023). Generative AI systems tend to perform well in lower-order cognitive tasks but continue to show limitations in higher-order domains such as Creating. This finding suggests the need to adapt Bloom's framework when applied to AI evaluation. Future adaptations may redefine creativity tasks as a combination of factual grounding, cultural synthesis, and controlled generative exploration rather than open-ended invention (Thanh et al., 2023). Traditionally, cultural synthesis refers to blending different cultural elements within a single literary work (Dulatkyzy, Kadisha, Nurmanova, & Nurgali, 2024). Cultural synthesis is also the purposeful bringing together multiple cultures into a joint social environment, often to affect social change (Harrison, 2024). Automated classification frameworks like AutoBloom have further illustrated how instructional materials can be systematically aligned with Bloom's Taxonomy, yielding actionable insights for curriculum design (Shaik et al., 2023). Building on these insights, we argue that future research should move toward hybrid models of Bloom's Taxonomy for AI evaluation, where lower-order tasks remain standardized while higher-order domains incorporate structured generative prompts and retrieval-anchored synthesis criteria. Such adaptations would enable Bloom's framework to reflect AI's evolving capabilities and constraints while retaining its pedagogical rigor (Lombardi, Podžaj, Maffei, & Traetta, 2024).

By integrating traditional cognitive levels with digital tools and activities, Bloom's Taxonomy has found further application and development in digital learning, enabling educators to effectively incorporate technological methods to enhance teaching quality and promote deep learning among students. This adaptability allows Bloom's Taxonomy to be relevant not only in traditional classrooms but also as a critical basis for assessing student learning outcomes in modern digital education. The concept of hybrid intelligence systems, combining human and artificial intelligence, has been explored (Dellermann et al., 2021). Earlier research also applied Bloom's Taxonomy to knowledge management systems (Rademacher, 1999). Bloom's Taxonomy divides the cognitive domain into six levels: remembering, understanding, applying, analyzing, evaluating, and creating. It provides educators with a systematic framework for designing curricula and assessing learning outcomes (Jiang et al., 2024; Z. Wang et al., 2024). Recent studies have explored the capabilities of large language models and multimodal LLMs across various domains using specialized benchmarks. These benchmarks assess models' performance in transportation, legal knowledge, and multimodal tasks (Fei et al., 2023; Zhang et al., 2024). Many of these evaluations are structured around cognitive frameworks like Bloom's Taxonomy, examining skills from basic recall to complex application. For instance, Transportation Games tests the first three levels of Bloom's Taxonomy in the transportation domain (Zhang et al., 2024), while MLLM-Bench covers all six levels for multimodal tasks . Law Bench assesses legal knowledge across three cognitive levels (Fei et al., 2023). These studies reveal both the impressive capabilities and limitations of current LLMs, highlighting areas for improvement in domain-specific applications. Although Bloom's Taxonomy has been extensively applied in educational assessment, its utilization within cultural evaluation remains markedly underexplored. The empirical evidence supporting its relevance to cultural or intercultural analysis is exceedingly limited, with only a single study directly addressing this dimension. The study examined the applicability of Bloom's hierarchical cognitive framework to intercultural communication in pluralistic societies, proposing that its six cognitive components may serve as analytical lenses for identifying moral and ethical parallels across distinct cultural systems, thereby facilitating a more nuanced understanding of cross-cultural interactions (Sharif & Shamsudin, 2017).

In the context of cultural knowledge, Bloom's Taxonomy can be effectively applied as an evaluative framework, particularly for assessing knowledge systems with rich historical backgrounds and profound cultural content, such as Hakka culture. At the first level, remembering, learners are expected to recall and recognize fundamental cultural facts, such as "What are the major traditional festivals of the Hakka group?" This level primarily tests the learner's grasp of basic cultural knowledge. The understanding level involves explaining and interpreting cultural content, as in the question "What are the characteristics of traditional Hakka architecture?" Here, learners are required not only to know the facts but also to understand the cultural significance behind them. The applying level further challenges learners to apply their cultural knowledge to new contexts, such as "How can Hakka architectural styles be incorporated into modern urban planning?" This requires learners to flexibly apply their knowledge. The analyzing level involves breaking down and comprehending cultural phenomena, for instance, "Analyze the structural differences between the Hakka language and other Chinese dialects." This demands that learners identify and compare relationships between different cultural elements. At the evaluating level, learners are expected to critically assess cultural phenomena, such as "Evaluate the impact of modernization on traditional Hakka culture," which requires independent thinking and judgment based on established criteria. Finally, the creating level represents the highest order of cognitive activity, where learners are expected to use their knowledge to create new cultural products or propose innovative cultural preservation strategies, such as "Design a modern media project to promote Hakka culture." This structured question-and-answer framework can comprehensively evaluate LLMs' cognitive abilities in handling Hakka cultural knowledge, thereby providing robust support for cultural preservation and education (Poornima, Kumar, & Ramesh, 2024; Spanos, 2024).

## 2.2. Retrieval-Augmented Generation

LLMs demonstrate remarkable capabilities but face challenges such as hallucinations, outdated information, and non-transparent, untraceable reasoning processes. RAG has emerged as a promising solution to address limitations of LLMs by incorporating external knowledge sources (Y. Gao et al., 2023; Huang & Huang, 2024). The fundamental concept of RAG involves combining knowledge retrieval with the text generation process, enabling the model not only to rely on its pre-trained internal knowledge to generate text but also to dynamically retrieve relevant information from external databases or knowledge repositories and incorporate this retrieved content into the final output. RAG enhances accuracy, credibility, and knowledge updating capabilities of LLMs, particularly for knowledge-intensive tasks (Lewis et al., 2020). The RAG paradigm typically comprises three core components: retrieval, generation, and augmentation, with multiple strategies available for integrating retrieved content into the generation process. Recent surveys have categorized RAG systems based on their architectural variations, outlined their technological foundations, and documented their advancements across different modalities and application domains (P. Zhao et al., 2024). RAG models have consistently demonstrated state-of-the-art performance on knowledge-intensive natural language processing (NLP) tasks, surpassing both parametric-only models and specialized task-specific architectures (Lewis et al., 2020). In the context of cultural archives, recent work exemplifies the application of NLP and data analytics to large-scale oral history collections (Chen, Kim, Chen, & Sakata, 2024). Cherukuri et al. (2025) proposed a scalable annotation framework combining expert curation, prompt engineering, and LLM-based semantic and sentiment analysis. Their findings show that with careful prompt design and retrieval augmentation, LLMs can achieve high semantic fidelity while preserving narrative authenticity and emotional nuance (Cherukuri, Moses, Sakata, Chen, & Chen, 2025). Building on this perspective, the present study extends such LLM methodologies to the Hakka cultural domain, emphasizing that culturally grounded benchmarks and retrieval-augmented frameworks are essential for mitigating AI hallucination risks and ensuring respectful representation of minority cultures. While RAG offers substantial accuracy and contextual relevance advantages, ongoing challenges remain in optimizing retrieval mechanisms, fusion strategies, and evaluation methods, signaling essential directions for future research and system enhancement.

## 2.3. Application of LLMs in Cultural Knowledge Assessment

The application of LLMs in cultural knowledge assessment, particularly in understanding and processing the knowledge of minority cultures, presents both challenges and opportunities. LLMs are trained on vast amounts of data, which enables them to perform effectively when handling mainstream cultural knowledge that is widely documented. However, when it comes to minority cultures, where such knowledge is often underrepresented in training datasets, LLMs face significant challenges in accurately understanding and generating content. The unique language, history, and customs of minority cultures may differ markedly from those of the mainstream, and in the absence of sufficient context or training examples, LLMs are prone to misunderstandings or the generation of incorrect information. Despite these challenges, there is also a substantial opportunity for LLMs to make meaningful contributions in these specialized areas, particularly with the application of enhanced techniques like RAG. By properly utilizing these techniques, LLMs can play a crucial role in the promotion and preservation of minority cultural knowledge, aiding in its transmission in the digital age and its dissemination in a globalized context.

Current methods for evaluating cultural knowledge in language models primarily rely on multitask testing and specific question-answering or summarization tasks, each with its own strengths and limitations. The Massive Multitask Language Understanding (MMLU) is a comprehensive multitask benchmark designed to assess LLMs' knowledge across various academic domains (Hendrycks et al., 2020). However, such benchmarks are often focused on academic knowledge, potentially offering limited insights into the nuanced understanding required for minority cultural knowledge. The Stanford Question Answering Dataset (SQuAD) tests LLMs' text comprehension through a question-and-answer format, making it suitable for evaluating the precision and contextual understanding of models when addressing specific queries (Rondeau & Hazen, 2018). XSum, an extreme summarization dataset, evaluates models' ability to condense and extract information, making it useful for assessing performance in generating concise and informative summaries (Narayan, Cohen, & Lapata, 2018). Nevertheless, it also faces the challenge of insufficient representation of minority cultural data in its training set. While these datasets provide a framework for evaluating LLMs' understanding of cultural knowledge, they have notable limitations, particularly regarding the detailed and accurate representation of minority cultures. Therefore, the development of specialized test datasets and evaluation methods tailored to specific cultural knowledge remains an important direction for future research.

Recent research has concentrated on creating comprehensive frameworks to evaluate LLMs across multiple dimensions. A multi-dimensional framework has been proposed to assess linguistic proficiency, task performance, ethical alignment, and societal impacts, providing a holistic approach to LLM evaluation (Noguer i Alonso, 2023). In the medical domain, COGNET-MD offers a specialized toolkit with varying difficulty levels in multiple-choice quizzes to rigorously evaluate LLMs' medical knowledge and application (Panagoulias et al., 2024). An alternative approach, focusing on language acquisition principles, advocates for shifting away from traditional metrics to foster interdisciplinary insights in LLM assessment (Vera, Moya, & Barraza, 2023). Another framework, FAC2E, dissociates language and cognitive capabilities, evaluating LLMs through a three-step process of recalling, utilizing, and applying knowledge to provide deeper insights into LLM performance (Wang, Wu, Ma, & Liu, 2024). Collectively, these frameworks aim to create nuanced assessments of LLMs, addressing critical challenges such as bias, fairness, and domain-specific performance, thereby supporting responsible development and deployment across fields like medicine and cognitive sciences.

3. **Research Methodology**

This study adopts a comprehensive evaluation framework that not only assesses the performance accuracy of LLMs but also critically examines their capability to handle societal, ethical, and fairness-related complexities inherent within diverse cultural contexts (Chang et al., 2024) . The evaluation model incorporates multiple dimensions, including representational equity, fairness in generated responses, and bias minimization, ensuring alignment with equitable principles across populations. To systematically evaluate LLMs' cognitive capabilities within the domain of minority cultural knowledge, specifically Taiwanese Hakka culture, this research leverages Bloom's Taxonomy as a structured approach. Bloom's cognitive domain encompasses hierarchical levels, including Remembering, Understanding, Applying, Analyzing, Evaluating, and Creating. Originally designed for educational curriculum and assessment, this taxonomy is adapted here to systematically measure the cognitive performance outcomes of LLMs, ranging from basic recall to advanced creative synthesis. An extensive dataset of Hakka cultural knowledge was compiled from diverse authoritative sources, including historical records, academic research, oral histories, and traditional practices, to establish a robust and representative question-answer dataset. This rich compilation addresses key cultural elements such as language, traditional architecture, culinary practices, festivals, and social customs, providing a diverse and detailed foundation for evaluating LLM performance. The evaluation dataset was structured according to Bloom's cognitive levels, with questions designed to progressively test the depth and breadth of the LLMs' comprehension. Questions range from foundational knowledge retrieval (e.g., "Identify major settlements of the Hakka group.") to complex creative scenarios (e.g., "Propose an innovative strategy to enhance global appreciation of Hakka culture."). By applying this hierarchical approach, the methodology systematically examines LLMs' understanding, analytical capabilities, and creative potential within a cultural context. This structured evaluation provides critical insights into the strategic integration of AI within digital archives, aiming to support cultural preservation and promote innovative modes of knowledge dissemination.

## 3.1. Research Design

The research design of this study is illustrated in Figure 1, which integrates RAG with Bloom's Taxonomy to examine how LLMs process and apply culturally specific knowledge. The framework establishes a stepwise process linking the creation of a cultural knowledge base, retrieval augmentation, cognitive evaluation, and model performance assessment.

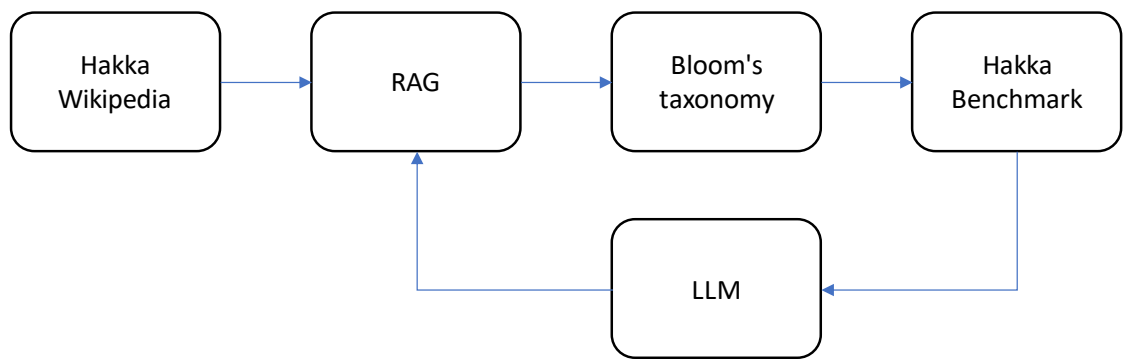

Figure 1. Research Model

The first stage focuses on building a domain-specific cultural knowledge base compiled from Hakka Wikipedia and other verified digital archives. This corpus encompasses linguistic, historical, and ethnographic materials that represent the language, customs, architecture, festivals, and oral heritage of the Hakka community. It provides the primary foundation for ensuring that all subsequent analyses are anchored in accurate and contextually meaningful cultural information. In the second stage, RAG functions as the connecting mechanism between the Hakka knowledge base and the LLM. During processing, RAG retrieves culturally relevant passages and integrates them into the model's reasoning process. This retrieval mechanism strengthens factual grounding, enhances contextual precision, and minimizes the risk of misrepresentation or cultural distortion in the generated responses. The third stage introduces Bloom's Taxonomy as a cognitive framework for organizing evaluation tasks. Each of the six cognitive domains is operationalized as a distinct layer of cultural cognition within the benchmark. These task categories collectively constitute the Hakka Cultural Benchmark, which serves as the empirical foundation for assessing the cognitive capabilities of LLMs. Each task is designed with clear objectives, verified answers, and standardized scoring criteria to maintain consistency and cultural relevance. In the final stage, the benchmark is used to evaluate LLM performance under two testing conditions. In the closed-book condition, models rely solely on internal knowledge, without access to external retrieval. In the open-book condition, RAG is activated to provide external cultural references. The results are analyzed through Bloom's framework to assess how effectively each model demonstrates factual accuracy, interpretive depth, and creative synthesis in relation to Hakka cultural content.

This research design offers a systematic and transparent methodology for evaluating the cultural competence of artificial intelligence systems. By combining the cognitive hierarchy of Bloom's Taxonomy with the contextual retrieval function of RAG, the framework enables an evidence-based assessment of how LLMs process, understand, and generate culturally grounded knowledge. Our system was designed to ensure accurate retrieval and contextual grounding of cultural information from the Hakka Cultural Knowledge Base. The process begins with text collection and preprocessing. Cultural texts were divided according to chapters and paragraphs, then further segmented into passage units of approximately 150 tokens. Each passage was converted into a semantic vector using an embedding model and stored in a vector database, along with its corresponding index and document identifiers, for efficient retrieval. During inference, the system performs both reasoning and retrieval operations. When a query is received, it first retrieves the top 30 most semantically similar passages ranked by cosine similarity. If the top three passages each exceed a similarity score of 70%, they are used directly as contextual references. Otherwise, a reranking process is applied to the top 30 passages, filtering those with similarity scores above 30% and selecting up to three of the most relevant. If the retrieved information is still insufficient, the system supplements the context by accessing the complete source document identified by its index. To ensure fairness and comparability, each model is allowed only one retrieval process per question under the RAG-enabled condition. The final responses are evaluated for factual correctness and cultural alignment. This configuration provides a balanced approach between retrieval precision and interpretive depth, allowing systematic comparison between closed-book and open-book (RAG-enabled) conditions while maintaining reproducibility and cultural fidelity.

### 3.2. Cognitive Domain Benchmark Construction

A benchmark refers to a standard or set of criteria against which the performance of different models can be compared (Sarkis, 2001; Talluri & Sarkis, 2001). In this study, a cognitive benchmarking dataset was constructed to assess LLMs' capabilities in understanding and applying Hakka cultural knowledge across six levels of Bloom's Taxonomy. The dataset consists entirely of multiple-choice questions with standard answers to ensure objectivity, replicability, and automated scoring.

This study adopts a systematic construction process to build a cognitive benchmarking dataset to enable a precise and structured evaluation of LLMs' capabilities in understanding Hakka cultural knowledge. The entire process is designed not only to assess LLMs' comprehension across various cognitive domains but also to test the effectiveness of RAG in improving cultural relevance and factual correctness. To ensure objective and quantifiable performance assessment, all questions in the dataset are designed as multiple-choice items with clearly defined standard answers. This approach allows for automated scoring and direct comparison across models, avoiding ambiguity in subjective interpretation and enhancing replicability in benchmarking tasks.

The construction process involved several stages designed to ensure both linguistic authenticity and cultural representativeness. First, a source corpus was compiled by collecting and digitizing materials from multiple Hakka cultural reference books and open-access online archives. These included a range of publications on Hakka customs, architecture, language, festivals, and oral heritage, which were consolidated into the Hakka Culture Encyclopedia containing 2,029 entries across 22 thematic domains. Each entry was standardized into an encyclopedic format that highlights key cultural concepts, terminology, and contextual explanations. Second, a complementary dataset was assembled from the Ministry of Education's Hakka Knowledge Base, which exists primarily as PDF documents on the official website. These files were converted into machine-readable text using OCR preprocessing and then segmented into dictionary-style entries through automated parsing scripts. Redundant, unclear, or overlapping information was removed to improve clarity and cultural coherence, resulting in a curated corpus of 1,693 usable entries. Third, based on this corpus, a prompt-driven question generation process was implemented to design cognitive evaluation items aligned with Bloom's six levels. Customized prompt templates were used to generate draft questions and answers for each entry, producing 36,522 initial items. These drafts were reviewed by domain experts in the Hakka language and culture to ensure conceptual validity, linguistic correctness, and cognitive appropriateness. Both the prompts and generated outputs were iteratively refined through several review cycles, during which experts identified and revised ambiguous formulations and culturally inaccurate interpretations. Fourth, a final quality control and expert validation stage produced a benchmark dataset of 10,158 culturally verified questions with corresponding answers.

The dataset items are meticulously designed to correspond with each level of Bloom's Taxonomy: (1) Remembering: Questions assess the model's ability to recall fundamental facts and definitions, such as "What are the main Hakka settlements located in Miaoli County?" (2) Understanding: Questions evaluate the comprehension of concepts and cultural phenomena, such as "What are the distinctive characteristics of traditional Hakka culinary practices?" (3) Applying: Questions test the application of cultural knowledge to new or practical contexts, such as "How can traditional Hakka clothing styles be incorporated into contemporary fashion designs?" (4) Analyzing: Questions require the examination and comparison of different cultural elements, such as "Compare and contrast Hakka opera with other forms of Han Chinese traditional theater." (5) Evaluating: Questions involve critical assessment and judgment of cultural developments and impacts, such as "Evaluate the historical evolution of the Hakka people's social status within Taiwanese society." (6) Creating: Questions challenge the model to generate innovative ideas and projects grounded in Hakka culture, such as "Design a modern art exhibition that effectively showcases and promotes Hakka cultural heritage."

### 3.3. Selection of Large Language Models

The selection of LLMs is a critical component of the experimental design. To ensure a comprehensive and representative evaluation of cultural knowledge processing, this study selected three widely used models: gpt-4.1-mini, gemini-2.5-flash, and llama-4-maverick, based on a multi-faceted rationale. These models were chosen to represent diversity across architectural designs, commercial openness, and integration capability with RAG frameworks, which aligns with the study's objective to benchmark AI performance across cognitive tasks involving cultural content (Bommasani, Liang, & Lee, 2023; Lewis et al., 2020). The four primary criteria for our selection were: (1) State-of-the-Art (SOTA) Performance, (2) Architectural and Philosophical Diversity,

(3) Relevance to Cultural Knowledge Representation, and (4) Efficacy in Retrieval-RAG frameworks (Mars, 2022; Tojima & Yoshida, 2025; Yu et al., 2023).

To reflect these criteria, the study incorporated three representative models. gpt-4.1-mini (OpenAI) exemplifies a commercial-grade system balancing cost and performance, while gemini-2.5-flash (Google) emphasizes rapid inference and efficient context handling. Together, they establish robust proprietary baselines. In contrast, llama-4-maverick (Meta) provides an open-source alternative that facilitates transparent access to weights and seamless RAG integration. This combination ensures that black-box commercial systems and community-driven models are represented, offering complementary perspectives on cultural knowledge tasks (Bommasani et al., 2023; Lewis et al., 2020).

Moreover, the three models differ in their training pipelines and alignment strategies, introducing variation in cultural content exposure and instruction-following behavior. Such diversity allows the benchmarking framework to capture how proprietary versus open-source development philosophies shape model performance in culturally grounded contexts. Finally, all selected models were tested with RAG integration: commercial APIs for gpt-4.1-mini and gemini-2.5-flash, and full pipeline customization for llama-4-maverick. This setup enables comparative evaluation across plug-and-play and fully controllable retrieval scenarios.

All selected models were evaluated for their compatibility with RAG frameworks, a core mechanism in our benchmarking process. While gpt-4.1-mini and gemini-2.5-flash are commercial APIs with strong instruction-tuned behavior, llama-4-maverick's open-source design allows complete RAG pipeline integration and custom retriever augmentation. This combination enables testing both the plug-and-play and fully controllable RAG scenarios across models, especially in high-context cultural domains that require grounding responses in retrieved knowledge sources.

### 3.4. Model Training and Testing

In the model training and testing phase, this study selected several advanced large language models, including GPT-4o, Claude, LLaMA3, and Gemini, and integrated LLaMA3 with RAG technology for training. The decision to integrate only LLaMA3 into the RAG framework stems from its relatively poor performance among the evaluated LLMs. By incorporating RAG with LLaMA3, this study aims to determine whether the RAG approach can enhance the model's understanding of Hakka cultural knowledge and improve its accuracy. RAG enhances the models' ability to process Hakka cultural knowledge by incorporating relevant information retrieved from external knowledge bases into the generation process. During training, each model underwent preliminary fine-tuning to equip it with foundational knowledge of Hakka culture. Subsequently, RAG technology was introduced, enabling the models to dynamically retrieve relevant content from external sources, such as the "Hakka Culture Encyclopedia," when answering questions. This approach allows the models to generate more accurate and comprehensive responses. The integration of RAG not only improved the models' performance in handling less common or highly specialized knowledge but also significantly enhanced their reasoning and generative capabilities.

Following the training process, the models were evaluated using the previously constructed dataset, which is structured according to the six levels of Bloom's Taxonomy: Remembering, Understanding, Applying, Analyzing, Evaluating, and Creating. The study conducted a comprehensive assessment of each model's performance across these cognitive levels, collecting performance metrics such as accuracy rate. These evaluation metrics provided a detailed analysis of the strengths and weaknesses of each model in understanding and applying Hakka cultural knowledge, offering valuable insights for further optimization of the models and enhancing the effectiveness of RAG technology in this domain.

## 4. Research Result
### 4.1. Prompt Design and Task Instructions

Two standardized prompt templates were developed to ensure consistency and comparability across experimental conditions. The first template corresponds to the open-book (RAG-enabled) condition, where the model can access the Hakka knowledge base for information retrieval. The second corresponds to the closed-book condition, where the model relies solely on its internal knowledge without external input. Both templates share the same task format, instructions, and response schema, differing only in the availability of retrieval access. Each prompt includes four components: task context and role definition, question set with the expected number of answers, answer format in JSON schema, and retrieval constraints. The final prompt structure was refined through iterative testing to ensure clarity, balanced cognitive difficulty, and cultural representativeness across Bloom's six cognitive domains. All items were reviewed by cultural and linguistic experts to verify their accuracy and relevance.

Agent prompts with tool:

*You are a participant in the Hakka Cultural Knowledge Competition. You are now taking an open-book exam, and you may use tools to search for information.*
*Please answer the following {expected_count} questions:*
*{formatted_questions}*

*# Notes*
*1. Please answer in JSON format as {"answer": [str, ...]}, only include the letter of the chosen option. For example: {"answer": ["A", "B", "C"]}.*
*2. You may take up to 10 reasoning steps, and must submit your final answers on the last step.*
*3. If the passages retrieved through vector_query are incomplete or insufficient, you may refer to the corresponding document ID to access the full content.*
*4. Only one retrieval (vector_query) is allowed per question.*
*5. Each answer must contain {expected_count} letters, with each being one of A, B, C, or D.*

Agent prompts without tool:

*You are a participant in the Hakka Cultural Knowledge Competition. You are now taking a closed-book exam, and you are not allowed to use any search tools.*
*Please answer the following {expected_count} questions:*
*{formatted_questions}*

*Please provide your final answer in the following format:*
*{"answer": ["A", "B", "C", "D", "A", "B"]}*

*Notes:*
*1. There are {expected_count} questions in total.*
*2. Provide only one letter (A, B, C, or D) per question.*
*3. Answer in the same order as the questions.*
*4. Do not use any external tools; rely solely on your internal knowledge.*
*5. Each answer must contain {expected_count} letters, with each being one of A, B, C, or D.*

**4.2. Evaluation Framework**

To systematically assess the cognitive capabilities of LLMs in processing Hakka cultural knowledge, this study adopts a structured evaluation framework grounded in Bloom's Taxonomy and supported by quantifiable performance metrics. All test items are designed as multiple-choice questions with clearly defined standard answers, enabling objective, automated evaluation of model responses. The primary evaluation metric used in this study is accuracy rate, calculated per model and cognitive domain using the following formula:

$$\text{Accuracy (\%)} = \frac{\text{Number of Correctly Answered Questions}}{\text{Total Number of Questions}}$$

This metric enables a consistent and interpretable comparison of model performance across Bloom's six cognitive domains: Remembering, Understanding, Applying, Analyzing, Evaluating, and Creating. Each domain emphasizes distinct cognitive functions and corresponds to different types of knowledge processing and reasoning: (1) Remembering: Measures the model's recall of factual cultural information (e.g., festivals, terminology, place names). Evaluation is based on direct answers that match standard facts. (2) Understanding: Assesses the model's semantic comprehension and its ability to accurately interpret culturally embedded meanings. The multiple-choice format evaluates whether the model can distinguish between closely related concepts and apply appropriate contextual understanding. (3) Applying: Evaluates the ability to transfer Hakka cultural knowledge to novel or practical contexts. Questions are scenario-based and require the model to infer correct applications (e.g., integration of traditional clothing in modern fashion). (4) Analyzing: Focuses on logical consistency, comparative reasoning, and the decomposition of cultural elements. Accuracy here reflects the model's ability to distinguish between nuanced relationships (e.g., comparing traditional theater forms). (5) Evaluating: Tests critical thinking and judgment based on cultural values or historical developments (e.g., the impact of modernization on Hakka practices). Options include varying degrees of relevance, and scoring favors culturally reasoned responses. (6) Creating: Although inherently generative, creative tasks were transformed into choice-based evaluative items. Expert panels predefined the most culturally appropriate and innovative answers, allowing the quantification of creativity in a culturally grounded way.

Experts designed and reviewed all questions to ensure alignment with cognitive complexity, cultural fidelity, and linguistic clarity. Using a multiple-choice format with standard answers enhances replicability and avoids ambiguity in scoring open-ended or generative responses. This design enables consistent and fine-grained evaluation of LLMs' cognitive processing of minority cultural knowledge. Applying this evaluation framework allowed for a comprehensive cross-model and cross-domain comparison, the results of which are presented in following section.

### 4.3. Data Analysis

To substantiate the observed performance differences among models and across cognitive domains, this study conducted a two-way Analysis of Variance (ANOVA) using model type and Bloom's Taxonomy category as independent factors, and accuracy rate as the dependent variable. The results revealed statistically significant differences between models ($F(5, 30) = 15.81$, $p < 0.000001$), indicating that the six LLMs demonstrated varying capabilities in processing cultural knowledge. However, the effect of the cognitive domain was not statistically significant at the 0.05 level ($F(6, 30) = 2.26$, $p = 0.065$), suggesting a more consistent performance pattern across different levels of Bloom's taxonomy. These results validate the appropriateness of Bloom's framework for LLM evaluation, highlighting that model selection plays a more critical role than task category in determining cultural knowledge accuracy.

Table 1 summarizes each model's accuracy and cost performance across cognitive domains. Cost (USD) represents the total evaluation costs incurred for each model during the full set of benchmark tasks. The analysis highlights key strengths and limitations of LLMs in processing cultural knowledge. In the Remembering domain, which evaluates the ability to recall factual Hakka cultural content (e.g., festivals, historical figures), the RAG-enhanced gemini-2.5 model achieved the highest accuracy (96.69%), far surpassing its non-RAG counterpart (76.55%). This underscores the essential role of retrieval-augmented mechanisms in enhancing factual recall, notably when base model pretraining lacks adequate minority cultural coverage. Both gpt-4.1 (RAG) and llama-4 (RAG) also showed strong performance (93.92% and 92.14%, respectively), suggesting robust integration of factual reinforcement.

In the Understanding domain, where models were required to interpret cultural meanings and contextual relationships, Gemini-2.5 (RAG) again led with 96.16%, followed closely by GPT-4.1 (RAG) at 94.57% and Llama-4 (RAG) at 94.27%. These results reflect the advantage of retrieval-enhanced architectures in grasping semantic depth and symbolic cultural patterns. Without RAG, model accuracy dropped by approximately 10%, revealing the limitations of standalone pretraining in capturing nuanced interpretations.

Application tasks evaluated the transferability of cultural knowledge to new or hypothetical scenarios. Here, gemini-2.5 (RAG) maintained its lead (92.44%), with gpt-4.1 (RAG) and llama-4 (RAG) closely trailing. The consistent RAG-driven improvement across models indicates that contextual document retrieval supports abstraction and adaptation of cultural principles to applied problem-solving.

Analysis tasks involved differentiating and organizing cultural phenomena, such as comparing Hakka traditions with other Han subcultures. The RAG-enhanced Gemini-2.5 model demonstrated top performance (92.50%), reflecting its capability in culturally grounded reasoning. While all RAG models showed improvements over their non-RAG versions, the gap in analytical accuracy was narrower than in other domains, suggesting some baseline capability in structural comparison even without retrieval.

In the Evaluation domain, the task assessed judgment and critical thinking regarding the authenticity, transformation, and value of cultural practices. The gemini-2.5 (RAG) model again achieved the highest performance at 91.61%, although the margin over other RAG models was modest. The similar performance across all RAG models suggests that evaluative tasks benefit from factual reinforcement. Still, it may also depend on the model's internal reasoning strategies, where architecture and fine-tuning play a role beyond retrieval.

Creation tasks presented the most significant challenge, requiring synthesis of cultural knowledge into novel outputs. Performance across models converged more closely here, with GPT-4.1 (RAG) achieving the highest score (86.00%), followed by Gemini-2.5 (RAG) and Llama-4 (RAG). The narrow spread suggests that while retrieval may support content accuracy, it constrains generative novelty. In contrast to tasks demanding recall or interpretation, creative generation requires models to move beyond retrieved content and form original ideas, where current RAG frameworks show inherent limitations.

Recent research explores the creative capabilities and limitations of LLMs across various domains. While LLMs can generate high-quality text and excel at stylistic reproduction, they often struggle with diversity, novelty, and originality in creative tasks (Ismayilzada, Stevenson, & van der Plas, 2024; Wenger & Kenett, 2025; Y. Zhao, Zhang, Li, & Li, 2025). The analysis confirms that RAG-enhanced models deliver superior performance across most cognitive domains, particularly in tasks emphasizing memory, comprehension, and application. However, in higher-order tasks like creative synthesis, the advantage of RAG diminishes. This pattern illustrates the importance of aligning model architectures and augmentation strategies with the cognitive demands of culturally grounded AI applications.

Table 1. Comparative Performance Metrics of LLMs across Cognitive Domain

| Model | llama-4 | gpt-4.1 | gemini-2.5 | llama-4(RAG) | gpt-4.1 (RAG) | gemini-2.5 (RAG) |
|---|---|---|---|---|---|---|
| 1. Memory | 69.34 | 68.81 | 76.55 | 92.14 | 93.92 | 96.69 |
| 2. Understanding | 84.23 | 84.52 | 86.18 | 94.27 | 94.57 | 96.16 |
| 3. Application | 79.33 | 80.09 | 82.34 | 90.14 | 90.49 | 92.44 |
| 4. Analysis | 80.57 | 82.34 | 81.57 | 90.19 | 89.55 | 92.50 |
| 5. Evaluation | 85.94 | 85.59 | 84.29 | 90.37 | 90.43 | 91.61 |
| 6. Creation | 83.34 | 84.82 | 82.75 | 85.65 | 86.00 | 85.76 |
| Overall Accuracy (%) | 80.46 | 81.03 | 82.28 | 90.46 | 90.82 | 92.53 |
| Cost (USD) | 11.40 | 6.56 | 36.19 | 11.40 | 6.56 | 36.19 |

Figure 2 illustrates the comparative accuracy of six LLMs across the six categories of Bloom's Taxonomy: Memory, Understanding, Application, Analysis, Evaluation, and Creation, based on their performance in Hakka cultural knowledge tasks. The radar chart reveals that RAG-enhanced models consistently outperform their non-RAG counterparts, particularly in lower and mid-level domains such as Memory and Understanding, with Gemini-2.5 (RAG) achieving the highest overall accuracy. This model is represented by a solid line to denote its superior performance. In contrast, baseline models exhibit flatter profiles, indicating more limited cognitive versatility. Notably, all models show smaller performance gaps in the Creation domain, suggesting that generative abstraction remains a common challenge regardless of retrieval integration. The grayscale line styles distinguish each model while maintaining visual clarity in a monochrome format. Radar chart showing model accuracy (%) across Bloom's six Bloom's Taxonomy cognitive domains. Axes are labeled from 60% to 100%. This chart highlights the importance of retrieval augmentation and architecture choice in optimizing AI systems for culturally grounded cognitive tasks.

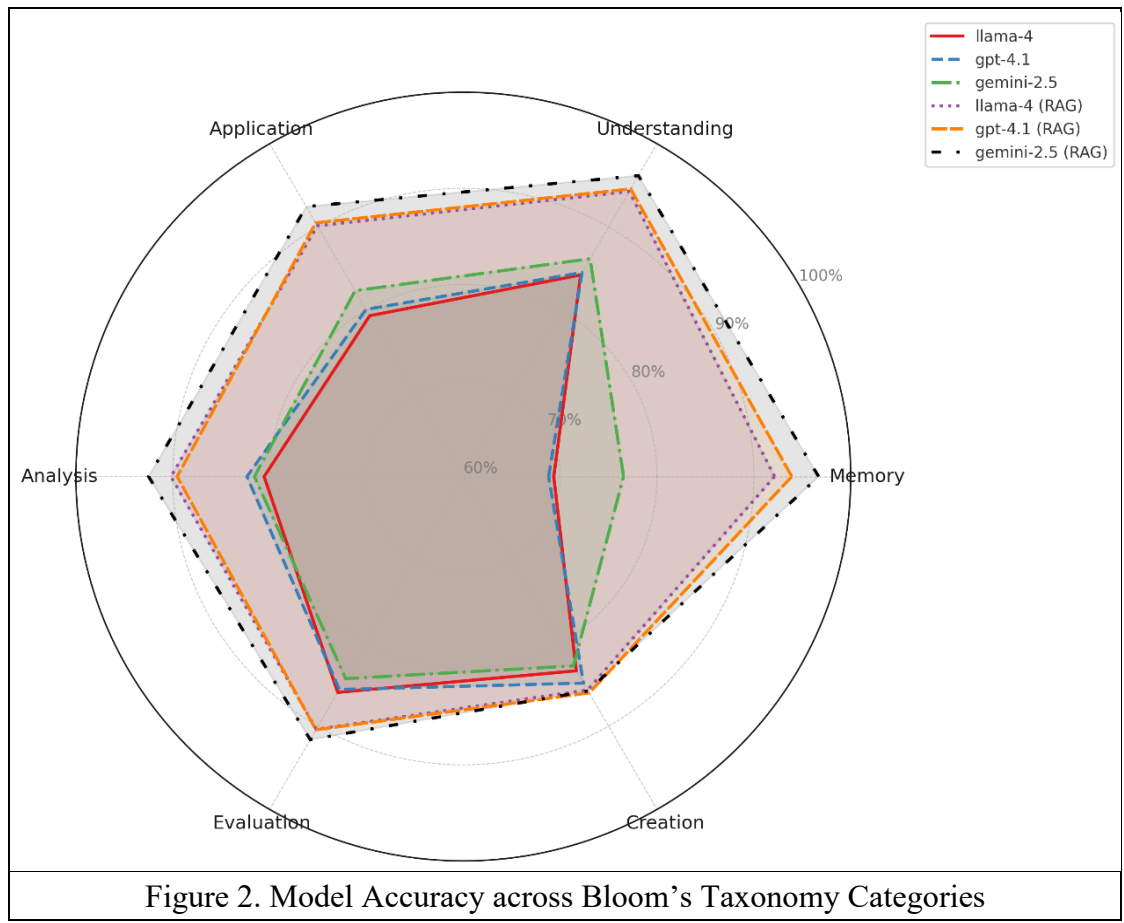

Figure 2. Model Accuracy across Bloom's Taxonomy Categories

## 4.4. Error Analysis of Model Performance

Based on the experimental results, the errors were classified into three main categories. The first category is content errors, which involve incorrect or culturally inaccurate facts. The second category is reasoning gaps, referring to flawed logic or misinterpretation when comparing or analyzing cultural elements. The third category is creativity deficiencies, which indicate a lack of originality or limited ability to generate culturally novel outputs. In the Memory category, non-RAG models such as LLaMA-4 and GPT-4.1 exhibited frequent content errors, including mislabeling traditional Hakka customs and confusing place-based identifiers. In contrast, RAG-enhanced models like Gemini-2.5 (RAG) showed significant improvements through external factual retrieval. In the Understanding and Application domains, reasoning gaps were prominent in baseline models. They often misunderstood cultural idioms or failed to adapt traditions in context. For example, models misaligned traditional Hakka practices with contemporary uses. RAG models, particularly Gemini-2.5 (RAG), mitigated many of these issues through contextual grounding, though some still struggled with nuanced interpretations. In Analysis and Evaluation, models such as LLaMA-4 and Gemini-2.5 (non-RAG) underperformed due to weak logical structuring and insufficient comparative insights, especially when distinguishing cultural subtleties among ethnic traditions. The Creating category exposed limitations in generative capacity. While GPT-4.1 and Gemini-2.5 (RAG) maintained relatively strong performance, even the top-performing models exhibited creativity deficiencies, often recycling factual content without introducing culturally innovative perspectives. Comparative error patterns revealed that RAG frameworks significantly reduced content-related mistakes in lower-order tasks but occasionally constrained originality due to overreliance on retrieved text. Conversely, non-RAG models displayed more stylistic fluency but lacked accuracy and contextual grounding. These findings suggest three major improvement strategies: expanding Hakka-specific corpora for model training and retrieval; fine-tuning LLMs on culturally rich, domain-specific prompts to strengthen contextual reasoning; and integrating hybrid strategies that blend retrieval grounding with generative flexibility. Error analysis thus underscores the dual challenge of factual precision and cultural creativity, offering a roadmap to enhance the cultural competence of LLMs for minority heritage preservation.

## 5. Discussion

### 5.1. Performance of LLMs Across Cognitive Domains

The experimental results reveal several critical insights into the capabilities and limitations of LLMs when tasked with processing minority cultural knowledge across varying cognitive demands. The consistent performance improvement observed in all RAG-augmented models across Bloom's Taxonomy levels affirms the strategic value of retrieval mechanisms in bolstering accuracy, contextual grounding, and semantic interpretation, particularly in underrepresented domains such as Hakka cultural knowledge. These findings align with existing literature on the benefits of retrieval augmentation for low-resource or culturally specific tasks, where internal model parameters may lack sufficient pretraining exposure.

Notably, the superiority of RAG-based models was evident in foundational cognitive domains such as Memory and Understanding and extended to higher-order tasks like Evaluation and Creation. Contrary to earlier assumptions that RAG architectures may restrict generative freedom, this study demonstrates that when supported by well-curated knowledge bases, RAG models can maintain or even enhance creative output while preserving factual integrity. This has significant implications for cultural heritage applications, where originality and authenticity are essential.

Among the evaluated models, Gemini-2.5 (RAG) consistently achieved the highest scores across all cognitive dimensions, suggesting that model architecture, training strategies, and retrieval integration collectively contribute to optimal performance. Furthermore, GPT-4.1 (RAG) displayed strong versatility, offering a favorable balance of cost efficiency and robust performance across tasks. Meanwhile, non-RAG models such as LLaMA-4 and GPT-4.1 (baseline) lagged in cultural knowledge recall and interpretation, underscoring the limitations of relying solely on static pretrained knowledge for specialized domains.

From a systems design perspective, these results highlight the need for context-aware AI configurations. For archival and cultural institutions aiming to deploy LLMs for preservation, education, or public engagement, model selection should align with task-specific cognitive demands. RAG-enhanced LLMs are especially suitable for ensuring accurate cultural transmission in memory-intensive or interpretive tasks, whereas additional tuning may be required to optimize creative outputs. Moreover, the relative performance consistency across models suggests that future research should prioritize refining retrieval content quality and cross-cultural evaluation frameworks, rather than focusing solely on architectural innovation.

This study validates Bloom's Taxonomy as a structured framework for evaluating cultural intelligence in LLMs. It advances the understanding of how retrieval-based techniques contribute to AI systems' accuracy, relevance, and adaptability in preserving minority cultures.

## 5.2. Application of Bloom's Taxonomy and Broader Implications

This study advances theoretical understanding by adapting Bloom's Taxonomy as an evaluative lens for assessing LLMs in cultural knowledge generation. Originally developed as a framework for educational assessment and curriculum design (Faraon, Granlund, & Rönkkö, 2023) , Bloom's hierarchical model is repurposed to diagnose and categorize the cognitive complexity of AI-generated content. Empirical results support prior findings that LLMs demonstrate higher accuracy and reliability in lower-order cognitive domains such as Remembering and Understanding, where tasks primarily involve recall or paraphrasing. However, performance diminishes in higher-order tasks such as Evaluating and Creating, where the models must demonstrate critical reasoning or cultural innovation. These limitations echo concerns that although LLMs can produce fluent and plausible outputs, they often lack the contextual grounding and interpretive coherence necessary for advanced cognitive tasks, particularly in culturally sensitive contexts (Yaacoub et al., 2025).

Beyond methodological contributions, the application of Bloom's framework highlights broader societal implications. In minority cultural preservation, structured evaluation helps prevent distortions and misrepresentations that could reinforce stereotypes or erode community heritage. By clarifying the strengths and weaknesses of LLMs, this approach informs the design of digital archives, museum resources, and educational tools that can equitably represent underrepresented cultures. The findings underscore that retrieval-augmented models enhance factual accuracy and promote representational equity, enabling culturally inclusive AI systems that support heritage transmission and knowledge access across generations. Moreover, Bloom's taxonomy provides policymakers and educators with a systematic basis to evaluate the reliability of AI-mediated cultural content, aligning digital governance with social responsibility.

The application of Bloom's framework also invites reflection on its epistemological assumptions. LLMs acquire knowledge through probabilistic associations rather than lived experience. As a result, certain cognitive dimensions, particularly those involving critical judgment or originality, may not directly translate from human cognition to machine cognition. This raises important questions about whether Bloom's Taxonomy, while useful as a comparative heuristic, might require recalibration when deployed in AI contexts. Nonetheless, its structured hierarchy offers a valuable tool for systematically evaluating the scope and limitations of LLM outputs, particularly in domains requiring cultural nuance, fairness in representation, and broader societal trust in AI systems.

## 6. Conclusion

### 6.1. Evaluating Cultural Knowledge Through Bloom's Taxonomy

This study contributes to the field of cultural knowledge management by applying Bloom's Taxonomy as a structured and replicable framework for assessing the cognitive performance of LLMs in minority culture representation. While LLMs have demonstrated strong performance in general language tasks, their training data often lacks sufficient depth and diversity at the cultural level, particularly regarding underrepresented communities such as the Taiwanese Hakka. This absence is not merely a matter of missing data points but reflects a systemic underrepresentation of culturally grounded narratives, epistemologies, and lived experiences. Simply supplementing training corpora with additional cultural texts may not fully resolve the problem, as foundational model architectures learn statistical correlations without the capacity for real-time contextual grounding. To address this, our research proposes the integration of RAG, which enables LLMs to access curated, community-authored knowledge dynamically during inference.

Experimental results show that RAG-enhanced models significantly outperform their non-RAG counterparts in lower to mid-order cognitive tasks, particularly in Remembering, Understanding, and Applying, because of their ability to retrieve relevant external information. However, even with RAG, performance in higher-order domains such as Creating remains limited. This finding suggests that cultural creativity and synthesis require more than factual retrieval and depend on interpretive sensitivity that current architectures still struggle to replicate. In the Creating domain, this study distinguishes between hallucination, which refers to the generation of culturally inaccurate or fabricated content, and cultural synthesis, which involves the constructive integration of verified cultural knowledge into novel outputs. While the former represents factual errors, the latter measures higher-order creative capacity. Empirical results show that models underperform for two reasons. They generate hallucinations. They also produce a synthesis that is insufficient or culturally biased. Both limit performance, but their implications for cultural evaluation differ. The framework classifies hallucination as a performance error. It treats successful cultural synthesis as a marker of advanced cognitive ability.

These findings emphasize the necessity of aligning LLM deployment strategies with cognitive task demands and highlight RAG as an efficient interim solution for enriching cultural accuracy until foundational models evolve to internalize better and represent culturally nuanced knowledge. Bloom's Taxonomy, in this context, proves to be an effective diagnostic tool for guiding the responsible application of LLMs in digital heritage and archival systems.

### 6.2. Enhancing Cultural Knowledge Retrieval with RAG

Incorporating RAG architecture has demonstrated marked improvements in the cultural fidelity, factual precision, and semantic depth of AI-generated content. When trained predominantly on general-purpose datasets, Standard LLMs frequently reflect structural biases and overlook the epistemic frameworks of underrepresented communities such as the Taiwanese Hakka. RAG addresses these limitations by enabling real-time access to curated, community-authored, and domain-relevant sources, anchoring outputs in more contextually accurate and culturally reflective knowledge bases. This retrieval mechanism enhances output quality regarding cultural alignment and contextual sensitivity and enables responsiveness to evolving or localized knowledge, an essential capability for heritage-focused applications. Additionally, RAG fosters greater representational equity by reducing overreliance on dominant cultural corpora, supporting initiatives in minority language revitalization, cultural preservation, and pedagogical content development. Looking forward, the advancement of RAG systems should emphasize integration with multimodal archival resources and deploying context-aware semantic retrieval methods to further enrich the inclusivity and applicability of AI in diverse cultural heritage ecosystems.

### 6.3. Implications for Cultural Preservation and Minority Heritage

The establishment of the Hakka Cultural Benchmark provides a systematic and replicable framework for evaluating how language models process, interpret, and generate culturally grounded knowledge. Beyond assessing existing large language models, this benchmark serves as an evaluative instrument for the development of culturally adaptive AI systems, including RAG pipelines and fine-tuned models designed for specific cultural contexts. It offers a measurable set of indicators that allow researchers and developers to quantify model performance across cognitive dimensions, ensuring that culturally sensitive reasoning and accurate contextual understanding are preserved throughout the AI development process. From a broader cultural preservation perspective, this framework represents a strategic response to the challenges faced by minority cultures in the age of artificial intelligence. As generative models increasingly influence how cultural narratives are created and transmitted, the risk of AI hallucination, which refers to the generation of inaccurate or decontextualized information, poses a serious threat to the authenticity and continuity of minority heritage. By grounding AI evaluation in a cognitively structured and culturally validated benchmark, this study demonstrates how computational systems can be directed toward cultural accountability. The benchmark not only enhances the interpretive transparency of AI but also enables communities to design, monitor, and refine language technologies that protect and revitalize their cultural knowledge in the digital era.

### 6.4. Research Limitation

While the proposed framework demonstrates promising results, several limitations require attention. The reliance on a domain-specific cultural dataset introduces potential bias. Representing cultural elements may not fully capture the diversity of real-world contexts. The choice of prompt design and task format also influences model performance. This effect is particularly evident in creative and open-ended cognitive domains. Evaluating higher-order skills, such as Creating, presents scalability challenges. Creativity is inherently subjective and requires nuanced human judgment, complicating large-scale assessment. The effectiveness of RAG-enhanced systems depends heavily on the quality of the retrieval corpus. Accuracy and contextual richness vary with the comprehensiveness of the underlying knowledge base. The selection of models was also constrained and did not cover the full spectrum of available LLM architectures. The RAG-enabled models in this study provide valuable insights into retrieval-augmented cultural knowledge generation. However, excluding other leading models, such as Claude or Mistral, limits generalizability. In addition, while commercial RAG integrations were used where available, the study did not investigate internal retrieval configurations or knowledge base differences. These factors may explain part of the observed performance variation. Addressing these limitations will require expanded model coverage. Iterative refinement of dataset construction and task design is also essential. Greater transparency in retrieval mechanisms is necessary to improve robustness and generalizability in future benchmarking studies. Another limitation concerns the design of tasks in the Creating domain. This study employed a multiple-choice format to ensure standardization and objective scoring across models. While this approach supports reliability and comparability, it inevitably restricts the expressive and generative range of the models and may therefore underestimate their creative potential. To address this limitation, future research should incorporate open-ended task formats that allow models to produce more diverse and contextually rich responses, thereby capturing a broader spectrum of creative reasoning. Furthermore, the inclusion of expert human evaluation mechanisms can provide qualitative assessments of originality, coherence, and cultural appropriateness in generated outputs. Such expert review processes would complement quantitative scoring and enhance the interpretive depth of creativity assessment. In addition, future studies may employ fine-tuned LLMs trained on culturally specific datasets as evaluators to automate the review process and reduce subjective bias. By combining standardized tasks with open-ended generation and expert or model-based evaluation, subsequent research can more accurately capture the higher-order cognitive and creative capacities of AI systems, especially in representing underrepresented cultures.

### 6.5. Future research

Future Research should broaden the scope of model evaluation to include a more diverse set of LLMs, particularly those with emerging or proprietary retrieval architectures. Comparative studies across commercial and open-source systems could help identify model-specific design factors influencing retrieval effectiveness in cultural tasks. In addition, further investigation into the quality and structure of the underlying knowledge sources used in RAG pipelines, including the relevance, diversity, and cultural fidelity of indexed documents, would help strengthen interpretive validity. Incorporating multilingual, multimodal, and community-curated datasets may also enhance RAG-based systems' inclusivity and cross-cultural adaptability. Future work can explore user-defined or domain-specific knowledge integration in real-time inference as APIs and developer tools evolve to support more configurable retrieval mechanisms. These directions will be essential for refining AI systems capable of generating dynamic, accurate, and culturally sensitive knowledge across global heritage contexts.